# Classification of jujube fruit based on several pricing factors using machine learning methods

Abdollah Zakeri
Faculty of Electrical and Computer engineering
University of Birjand
Birjand, Iran
a.zakeri@birjand.ac.ir

Ruhollah Hedayati
Computer engineering department
Isfahan University
Isfahan, Iran
roh.hedayati@eng.ui.ac.ir

Mohammad Khedmati
Faculty of Electrical and Computer engineering
University of Birjand
Birjand, Iran
mkhedmati@birjand.ac.ir

Mehran Taghipour-Gorjikolaie
Faculty of Electrical and Computer engineering
University of Birjand
Birjand, Iran
mtaghipour@birjand.ac.ir

*Abstract*—Jujube is a fruit mainly cultivated in India, China and Iran and has many health benefits. It is sold both fresh and dried. There are several factors in jujube pricing such as weight, wrinkles and defections. Some jujube farmers sell their product all at once, without any proper sorting or classification, for an average price. Our studies and experiences show that their profit can increase significantly if their product is sold after the sorting process. There are some traditional sorting methods for dried jujube fruit but they are costly, time consuming and can be inaccurate due to human error. Nowadays, computer vision combined with machine learning methods, is used increasingly in food industry for sorting and classification purposes and solve many of the traditional sorting methods' problems. In this paper we are proposing a computer vision-based method for grading jujube fruits using machine learning techniques which will take most of the important pricing factors into account and can be used to increase the profit of farmers. In this method we first acquire several images from different samples and then extract their visual features such as color features, shape and size features, texture features, defection and wrinkle features and then we select the most useful features using feature selection algorithms like PCA and CFS. A feature vector is obtained for each sample and we use these vectors to train our classifiers to be able to specify the corresponding pre-defined group for each of the samples. We used different classifiers and training methods in order to obtain the best result and by using decision tree we could reach 98.8% accuracy of the classification.

## I. Introduction

Ziziphus jujube, commonly called jujube (also known as red date, Chinese date, Korean date, Indian date) is a species of Ziziphus in the buckthorn family. It is mainly cultivated in southern Asia, northern India, southern and central regions of China, eastern and central regions of Iran. Jujube has many health benefits such as treating cancer, improving sleep and treating insomnia, improving heart health and decreasing the risk of heart disease. Jujube is sold both fresh and dried, but the fresh jujube is available only in certain months (from July until last of November) and due to its decrease of weight in the process of drying, some farmers prefer to sell their product when it is fresh.

There are several factors for jujube pricing. One of the most important factors, as mentioned before, is the weight of jujube. Size and shape specifications such as diameter and volume are also important for jujube pricing. Surface texture characteristics such as wrinkles and defections are important for most of buyers as well.

Most of jujube farmers in Iran sell their product after the process of drying while they only involve the total weight factor in pricing and consider an average price for all products setting aside all size and quality features. Our studies show that their profit could be increased significantly depending on the average quality of the product. Some farmers use traditional methods of sorting their products before selling, but these methods mainly focus on the size of jujube, and other pricing factors mentioned above are not considered in them. These methods are also time consuming, costly and can be inaccurate due to human error.

Nowadays, computer vision and artificial intelligence techniques are used increasingly in food industry for several purposes such as automated classification of agricultural products and fruits. Most of jujube grading and classification systems mainly focus on features such as shape, size and defections [1-4]. Wrinkle based classification of jujube fruit can be done by a combination of different transforms, morphological operations and segmentation methods.

Wang [5] extracted the wrinkle features on the surface of a jujube using a wavelet transform combined with morphological operations. But shallow wrinkles were not detected and the accuracy of this methodology is not enough to be used as a feature for a grading system.

Zhang et al. [6] used morphological operations, minima imposition method and watershed segmentation to detect wrinkles using a dataset of 304 jujube fruits and obtained 92.11% accuracy in grading them. However other features such as shape, size and defections were not considered.

In this paper we are proposing a computer vision-based method for grading jujube fruits using machine learning techniques which will take most of the important pricing factors into account and can be used to increase the profit of farmers.

## II. MODEL AND METHODOLOGY

The method proposed in this paper includes image pre-processing actions such as noise removal and shadow removal. Image segmentation techniques were used in order to subtract background from the sample image and then cropping the sample image to jujube's bounding box. The cropped image was used to obtain the feature vector including size and shape features, color features, texture features, wrinkle features and defection feature. The feature vector was used in different data mining and machine learning algorithms and in order to increase the accuracy of this algorithms, feature selection was done on this vector in order to extract the most useful features. Same algorithms were performed on the selected features vector and they showed significant increase in classification accuracy.

The overall structure of the proposed method is presented in the following illustration:

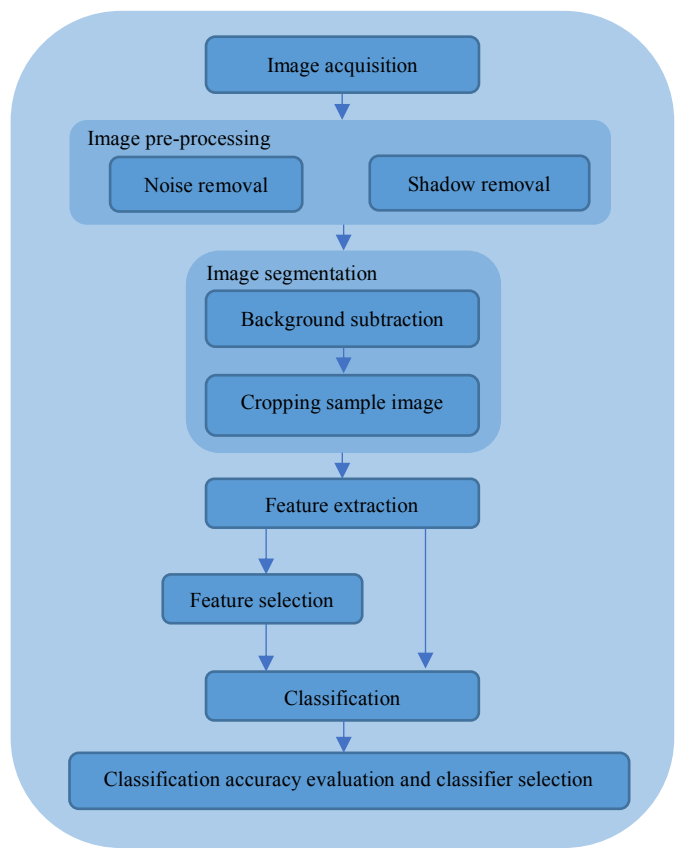

Fig. 1. Overall structure of proposed method

### A. Image acquisition

A machine vision system was developed to acquire images of jujube fruit (Fig 2).

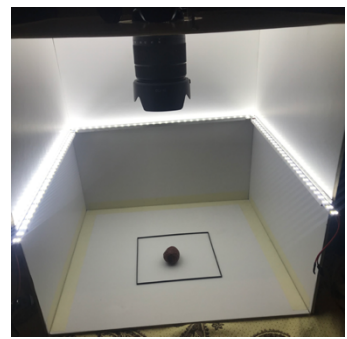

Fig. 2. Image aquisision system

This system consists of three SMD light sets, one on each side for propper illumination purposes and shadow removal, a white colored background that helps us subtract the background and segment the sample image, and a physical bounding box (15*15 cm) which will help us convert pixel values to millimeter or centimeter.

To acquire images, we used Cannon 80D with a Ba 18-135 optical lens 30 centimeters above the sample and captured images were 4000 by 6000 pixels.

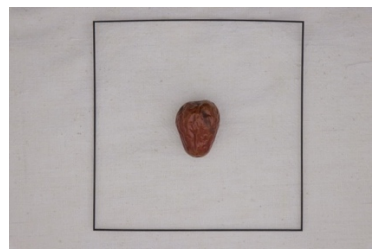

Fig. 3. Sample captured image of computer vision system

### B. Image pre-processing

Initially, the RGB image (fig 4.a) is converted to grayscale (fig 4.b), then the grayscale image is converted to binary image using Otsu's threshold (fig 4.c), next the original image was cropped using the physical bounding box region extracted from binary image (fig 4.d).

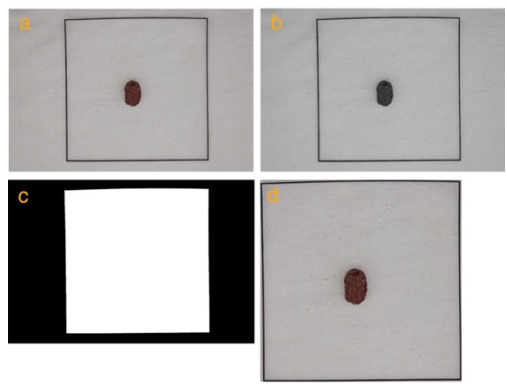

Fig. 4. Image pre-processing steps

## C. Image segmentation

In this section we subtracted the background from the sample image and we cropped the sample image to the jujube using following steps:

a) Binarizing the input image (fig 4.d) with Otsu's threshold, combined with morphological operations (fig 5.b).

b) Subtracting the background from input image (fig 4.d) by applying color thresholding methods (fig 5.c).

c) Cropping the background-less image (fig 5.c) to the sample's region extracted from binary image (fig 5.b) which results in separated jujube image (fig 5.d).

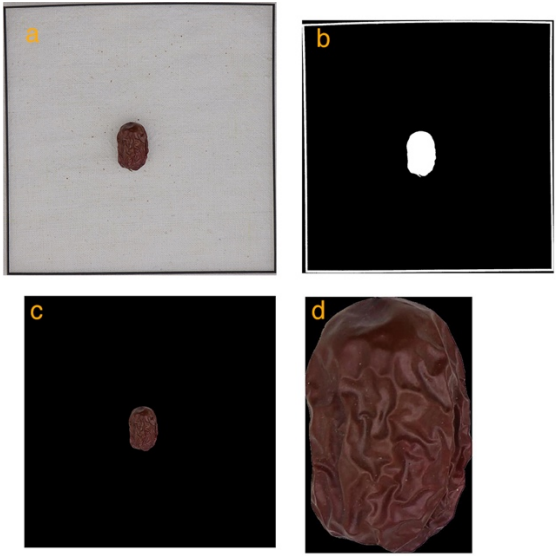

Fig. 5. Image segmentation steps

## D. Feature extraction

According to various studies conducted on object identification and the classification of different fruits, several features were used including texture, wrinkle and defection, shape, size features and color properties. Further details are provided on each of these features.

## E. Feature selection

In this section, we selected more useful features using feature selection algorithms such as PCA and CFS and the results indicate great improvements in classification.

## F. Classification

The last section of the method consists of grading sample images using the vector of features obtained in previous sections and classifying samples according to overall grade of each one. The aim of this section is designing an automatic procedure which classifies each sample in its corresponding pre-defined class using its vector of features.

We divided our sample data randomly into three parts:

a) Data used for training the classifier (70%)

b) Validation data used to prevent model over-fitting and affiliation of model to training data (15%)

c) Test data used for Model's accuracy evaluation purposes (15%)

## III. FEATURE EXTRACTION

### A. Shape and size features

Following shape and size features where extracted using functions provided in Matlab image processing toolbox.

TABLE I. SHAPE AND SIZE FEATURES

| ID | Title | Description |
|---|---|---|
| F1 | Area | Actual number of pixels in the region, returned in millimeters. |
| F2 | Perimeter | Distance around the boundary of the region. returned as a scalar. |
| F3 | Major axis length | Length (in pixels) of the major axis of the ellipse that has the same normalized second central moments as the region, returned as a scalar. |
| F4 | Minor axis length | Length (in pixels) of the minor axis of the ellipse that has the same normalized second central moments as the region, returned as a scalar. |
| F5 | Equivalent diameter | Equivalent diameter length of the jujube in millimeters |
| F6 | Solidity | Proportion of the pixels in the convex hull that are also in the region, returned as a scalar. |
| F7 | Eccentricity | Eccentricity of the ellipse that has the same second-moments as the region, returned as a scalar. |

a. The descriptions are extracted from Matlab documentations

All of features in *Table I* which are measured in millimeters where first extracted in pixels and then converted to millimeters using the size ratio of a fixed-size physical bounding box placed in all sample images.

## B. Color features

Several statistical color features where measured in 4 different color spaces and are listed in table below.

TABLE II.    SHAPE AND SIZE FEATURES

| Color space | Formula | Value | ID |
|---|---|---|---|
| RGB | Mean: $\mu = \frac{\sum_{x=0}^{z-1} h(x)}{z}$ | R | F8 |
| | | G | F9 |
| | | B | F10 |
| | | R / (R+G+B) | F11 |
| | | G / (R+G+B) | F12 |
| | | B / (R+G+B) | F13 |
| | | R-G | F14 |
| | | G-B | F15 |
| | | R-B | F16 |
| | Variance: $\sigma = \frac{\sum_{x=0}^{z-1}(h(x)-\mu)^2}{z}$ | R | F17 |
| | | G | F18 |
| | | B | F19 |
| | | R / (R+G+B) | F20 |
| | | G / (R+G+B) | F21 |
| | | B / (R+G+B) | F22 |
| | | R-G | F23 |
| | | G-B | F24 |
| | | R-B | F25 |
| RGB | Skewness: $s = \frac{\sum_{x=0}^{z-1}(h(x)-\mu)^3}{z\sigma^3}$ | R | F8 |
| | | G | F26 |
| | | B | F27 |
| | | R / (R+G+B) | F28 |
| | | G / (R+G+B) | F29 |
| | | B / (R+G+B) | F30 |
| | | R-G | F31 |
| | | G-B | F32 |
| | | R-B | F33 |
| | Kurtosis: $k = \frac{\sum_{x=0}^{z-1}(h(x)-\mu)^4}{z\sigma^4} - 3$ | R | F34 |
| | | G | F35 |
| | | B | F36 |
| | | R / (R+G+B) | F37 |
| | | G / (R+G+B) | F38 |
| | | B / (R+G+B) | F39 |
| | | R-G | F40 |
| | | G-B | F41 |
| | | R-B | F42 |
| L*a*b* | Mean: $\mu = \frac{\sum_{x=0}^{z-1} h(x)}{z}$ | L* | F43 |
| | - | a* | F46 |
| | | b* | F45 |
| HSV | Mean: $\mu = \frac{\sum_{x=0}^{z-1} h(x)}{z}$ | H | F46 |
| | | S | F47 |
| | | V | F48 |
| YCbCr | Mean: $\mu = \frac{\sum_{x=0}^{z-1} h(x)}{z}$ | Y | F49 |
| | | Cb | F50 |
| | | Cr | F51 |

a. h(x) is the grey level of pixels in the image with a pixel position of x , x can take any value between 1 and $z = m \times n$, where m and n are number of rows and columns of the image matrix, respectively.

b. μ, σ, S, and K are the Mean, Variance, Skewness, and Kurtosis of image pixels, respectively.

## C. Texture features

These features where calculated using GCLM (gray-level co-occurrence matrix).

TABLE III.    TEXTURE FEATURES

| Title | Description | Formula | ID |
|---|---|---|---|
| Contrast | Returns a measure of the intensity contrast between a pixel and its neighbor over the whole image. | $\sum_{i,j} \|i - j\|^2 p(i,j)$ | F52 |
| Correlation | Returns a measure of how correlated a pixel is to its neighbor over the whole image. | $\sum_{i,j} \frac{(i - \mu i)(j - \mu j)p(i,j)}{\sigma_i \sigma_j}$ | F53 |
| Energy | Returns the sum of squared elements in the GLCM. | $\sum_{i,j} p(i,j)^2$ | F54 |
| Homogeneity | Returns a value that measures the closeness of the distribution of elements in the GLCM to the GLCM diagonal. | $\sum_{i,j} \frac{p(i,j)}{1 + \|i - j\|}$ | F55 |

## D. Wrinkle and defection features

Due to color differences between jujube surface and defects on it, we were able to calculate the amount of defections by using color thresholding methods, then calculating area of the defected zone in pixels, converting it to square millimeters and dividing it by the jujube area (F1) gave us the defect proportion that we used as defection feature (F56).

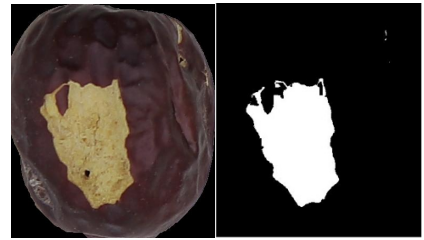

Fig. 6.    Jujube defection feature

For the wrinkle features, we used method explained in Zhang et al.[6] consisting of morphological operations and watershed segmentation to separate the wrinkle areas, count the total number of wrinkles on the jujube surface (F57) and total wrinkle area to jujube area ratio (F58)

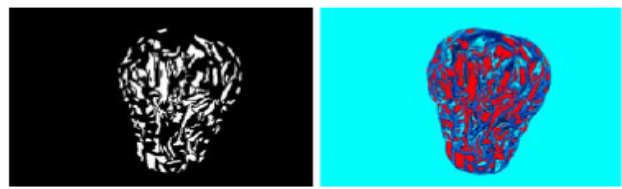

Fig. 7.    Wrinkle feature extraction using watershed segmentation

## IV. FEATURE SELECTION

In this section we used several feature selection methods including PCA and CFS to separate more useful features and classifiers showed significant improvement.

Approaches to feature transformation include:

1) Principal component analysis (PCA), used to summarize data in fewer dimensions by projection onto a unique orthogonal basis
2) Factor analysis, used to build explanatory models of data correlations
3) Nonnegative matrix factorization, used when model terms must represent non-negative quantities, such as physical quantities
4) Correlation based Feature Selection (PCA) is an algorithm that couples this evaluation formula with an appropriate correlation measure and a heuristic search strategy

The result of performing the above algorithms showed that the features list below are the most useful.

## V. CLASSIFICATION

The purpose of this section is to design an automatic procedure that specifies the corresponding pre-defined group for each sample image, using it's feature vector obtained in previous sections. The description of each of the techniques used is as follows.

### A. Artificial Neural Networks (ANN)

A neural network is a computing model whose layered structure resembles the networked structure of neurons in the brain, with layers of connected nodes. A neural network can learn from data, so it can be trained to recognize patterns, classify data, and forecast future events.

By designing an artificial neural network we tried to classify jujube samples into corresponding groups. The steps involved in this routine are as followed.

a) Create the neural network

b) Configure the network's inputs and outputs

c) Tune the network parameters (the weights and biases) to optimize performance

d) Train the network

e) Validate the network's results

We trained the neural network with different methods such as Levenberg-Marquardt and Bayesian Regularization

### B. Decision Tree

Decision tree learning is a method commonly used in data mining. The goal is to create a model that predicts the value of a target variable based on several input variables.
Some advantages of decision trees are:

- Simple to understand and to interpret. Trees can be visualized.

- Requires little data preparation. Other techniques often require data normalization. Note however that this module does not support missing values.

- The cost of using the tree (i.e., predicting data) is logarithmic in the number of data points used to train the tree.

- Able to handle both numerical and categorical data. Other techniques are usually specialized in analyzing datasets that have only one type of variable.

- Able to handle multi-output problems.

- Uses a white box model. If a given situation is observable in a model, the explanation for the condition is easily explained by Boolean logic. By contrast, in a black box model (e.g., in an artificial neural network), results may be more difficult to interpret.

- Possible to validate a model using statistical tests. That makes it possible to account for the reliability of the model.

### C. Support Vector Machine (SVM)

A Support Vector Machine (SVM) is a discriminative classifier formally defined by a separating hyperplane. In other words, given labeled training data (supervised learning), the algorithm outputs an optimal hyperplane which categorizes new examples. In two dimensional space this hyperplane is a line dividing a plane in two parts where in each class lay in either side.

Several kernel functions including second and third degree polynomial and Gaussian were used and obtained results were compared with each other.

### D. K-Nearest Neighbour (k-NN)

K-NN is a non-parametric method used for classification and regression. In this method, an object is classified by a plurality vote of its neighbors, with the object being assigned to the class most common among its k nearest neighbors.
We trained several models by changing the value of k and weighing our data.

## VI. TESTS AND RESULTS

We divided our data into three different groups in order to train the neural network:

a) Training data (70%) which is used to train the classifier

b) Validation data (15%) used to avoid over-fitting and lower the dependecy of the model on training data

c) Test data (15%) used to evaluate the accuracy of the trained model.

In other learning methods we used 10-fold cross-validation. Initially, the training process was performed with the default values of the parameters and then the parameters were changed to improve the methods used. In the following, the results obtained from each method are presented before and after applying the feature selection algorithms.

TABLE IV. MODELS' ACCURACY RESULTS

| Model | Training method | ACC(%) before FS | ACC(%) after FS |
|---|---|---|---|
| ANN | Levenberg-Marquardt | 92.7 | 96.7 |
| ANN | Bayesian Regularization | 91.5 | 91.7 |
| SVM | Linear kernel function | 82 | 85.7 |
| SVM | Second-degree kernel function | 82.3 | 82.5 |
| SVM | Third-degree kernel function | 80.1 | 82 |
| SVM | Gaussian kernel function | 77.1 | 81.3 |
| Decision tree | Preset: Simple Tree<br>Maximum number of splits: 4<br>Split criterion: Gini's diversity index | 91.4 | 93.2 |
| Decision tree | Preset: Medium Tree<br>Maximum number of splits: 20<br>Split criterion: Gini's diversity index | 98.2 | 98.8 |
| Decision tree | Preset: Complex Tree<br>Maximum number of splits: 100<br>Split criterion: Gini's diversity index | 98.2 | 98.7 |
| KNN | Whighted KNN<br>Number of neighbors: 10<br>Distance metric: Euclidean<br>Distance weight: Squared inverse | 70 | 75.6 |
| KNN | Preset: Cosine KNN<br>Number of neighbors: 10<br>Distance metric: Cosine | 68.2 | 70.3 |
| KNN | Preset: Cubic KNN<br>Number of neighbors: 10<br>Distance metric: Minkowski (cubic) | 67 | 67.1 |